\documentclass[sigconf]{acmart}

\AtBeginDocument{%
  \providecommand\BibTeX{{%
    \normalfont B\kern-0.5em{\scshape i\kern-0.25em b}\kern-0.8em\TeX}}}

\copyrightyear{2023} 
\acmYear{2023} 
\setcopyright{acmlicensed}
\acmConference[MM '23]{Proceedings of the 31st ACM International Conference on Multimedia}{October 29-November 3, 2023}{Ottawa, ON, Canada}
\acmBooktitle{Proceedings of the 31st ACM International Conference on Multimedia (MM '23), October 29-November 3, 2023, Ottawa, ON, Canada}
\acmPrice{15.00}
\acmDOI{10.1145/3581783.3612142}
\acmISBN{979-8-4007-0108-5/23/10}

\usepackage{booktabs}                                   
\usepackage{multirow}                                   
\usepackage{tablefootnote}                              
\usepackage[symbol]{footmisc}                           
\usepackage{xcolor}                                     
\usepackage{enumitem}                                   
\usepackage{marvosym}                                   
\usepackage{ifsym}                                      
\usepackage{stfloats}                                   

\newcommand{\para}[1]{\subsubsection{#1}}

\newcommand{\eat}[1]{}                                  

\def\dtf{DTF-Net}

\begin{document}


\title[{\dtf}: Category-Level Pose Estimation and Shape Reconstruction via Deformable Template Field]{{\dtf}: Category-Level Pose Estimation and\\Shape Reconstruction via Deformable Template Field}

\author{Haowen Wang}
\authornote{Equal Contributions. Work done when Haowen Wang interned at Midea Group.}
\author{Zhipeng Fan}
\authornotemark[1]
\email{hw.wang@bupt.edu.cn}
\email{fzp@bupt.edu.cn}
\affiliation{%
  \institution{
      State Key Laboratory of Networking and Switching Technology\\
      Beijing University of Posts and Telecommunications
  }
  \city{Beijing}
  \country{China}
}

\author{Zhen Zhao} 
\author{Zhengping Che} 
\author{Zhiyuan Xu} 
\email{zhaozhen8@midea.com}
\email{chezp@midea.com}
\email{chezhengping@gmail.com}
\email{xuzy70@midea.com}
\affiliation{%
  \institution{Midea Group}
  \city{Beijing}
  \country{China}
}

\author{Dong Liu} 
\author{Feifei Feng} 
\email{liudong13@midea.com}
\email{feifei.feng@midea.com}
\affiliation{%
  \institution{Midea Group}
  \city{Shanghai}
  \country{China}
}

\author{Yakun Huang} 
\email{ykhuang@bupt.edu.cn}
\affiliation{%
  \institution{
      State Key Laboratory of Networking and Switching Technology\\
      Beijing University of Posts and Telecommunications
  }
  \city{Beijing}
  \country{China}
}

\author{Xiuquan Qiao}
\authornote{Xiuquan Qiao and Jian Tang are the corresponding authors.}
\email{qiaoxq@bupt.edu.cn} 
\affiliation{%
  \institution{
      State Key Laboratory of Networking and Switching Technology\\
      Beijing University of Posts and Telecommunications
  }
  \city{Beijing}
  \country{China}
}

\author{Jian Tang} 
\authornotemark[2]
\email{tangjian22@midea.com}
\affiliation{%
  \institution{Midea Group}
  \city{Beijing}
  \country{China}
}

\renewcommand{\shortauthors}{Haowen Wang et al.}

\begin{abstract}
    Estimating 6D poses and reconstructing 3D shapes of objects in open-world scenes from RGB-depth image pairs is challenging.
Many existing methods rely on learning geometric features that correspond to specific templates while disregarding shape variations and pose differences among objects in the same category.
As a result, these methods underperform when handling unseen object instances in complex environments.
In contrast, other approaches aim to achieve category-level estimation and reconstruction by leveraging normalized geometric structure priors,
but the static prior-based reconstruction struggles with substantial intra-class variations.
To solve these problems, we propose the {\dtf}, a novel framework for pose estimation and shape reconstruction based on implicit neural fields of object categories.
In {\dtf}, we design a deformable template field to represent the general category-wise shape latent features and intra-category geometric deformation features.
The field establishes continuous shape correspondences, deforming the category template into arbitrary observed instances to accomplish shape reconstruction.
We introduce a pose regression module that shares the deformation features and template codes from the fields to estimate the accurate 6D pose of each object in the scene.
We integrate a multi-modal representation extraction module to extract object features and semantic masks, enabling end-to-end inference.
Moreover, during training, we implement a shape-invariant training strategy and a viewpoint sampling method to further enhance the model's capability to extract object pose features.
Extensive experiments on the REAL275 and CAMERA25 datasets demonstrate the superiority of {\dtf} in both synthetic and real scenes.
Furthermore, we show that {\dtf} effectively supports grasping tasks with a real robot arm.

\end{abstract}


\maketitle

\section{Introduction}
Object 6D pose estimation (i.e., predicting the 3D position and 3D orientation in the world coordinate system) and shape reconstruction (i.e., recovering the 3D geometric structure of objects from observed visual representations) are fundamental for many multi-media applications, such as augmented reality~\cite{marchand2015pose,rambach20186dof,su2019deep,wang2021geopose}, virtual reality~\cite{sra2016procedurally,lindlbauer2017changing}, scene understanding~\cite{sui2017sum,zhang2021holistic}, and robotic manipulation~\cite{deng2020self,wen2021gccn}.
Many researchers have explored this domain with the advance of deep learning and obtained huge progress in recent years~\cite{rad2017bb8,kehl2017ssd,oberweger2018making}.
Early methods~\cite{manhardt2018deep,li2018deepim,park2019pix2pose} rely on instance-level CAD models and point-to-point correspondences to perform pose estimations and reconstructions,
but the prerequisite of high-quality CAD models hampers the applications of these models to unseen objects without available 3D models in real-world scenarios.
Therefore, it is more desirable but challenging to conduct category-level estimation and reconstruction, i.e., utilizing categorical but not instance-level prior knowledge to estimate and reconstruct objects.

\begin{figure}[!t]
  \centering
  \includegraphics[width=.9\linewidth]{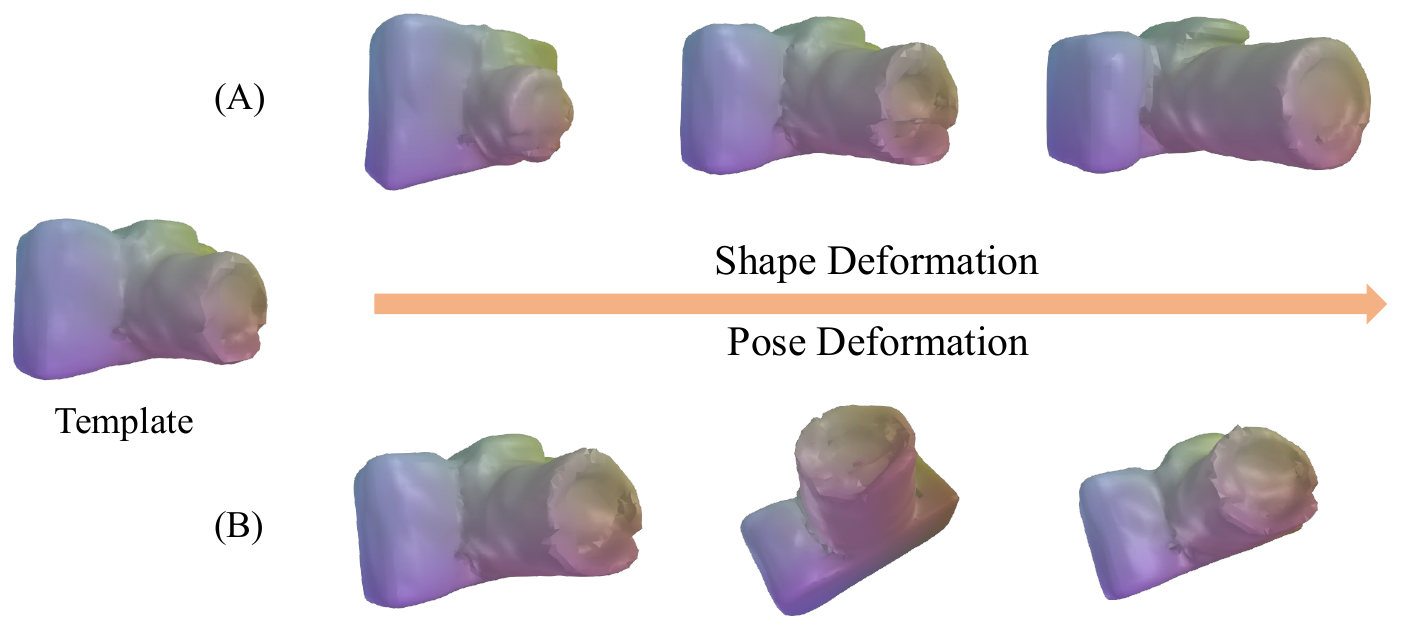}
  \caption{
  Illustrations of the two deformation types considered in {\dtf}.
  Given a categorical object template,
  (A) shape deformation refers to transforming the object's shape into any form,
  and (B) the pose deformation refers to generating the object in any spatial pose.
  By decoupling the two deformations, {\dtf} improves the capacity of both pose estimation and shape reconstruction.
  }
  \label{fig:deformation}
\end{figure}

Through the path of category-level estimation and reconstruction,
recent studies~\cite{tian2020shape,chen2020learning,deng2021deformed,lin2022sar,irshad2022centersnap} have concentrated on analyzing the common features of objects within the same category. Some methods~\cite{tian2020shape,lee2021category,zou20226d} employ Normalized Object Coordinate Space~\cite{wang2019normalized} to obtain a category shape prior by unifying different instances, and they estimate the 6D pose by calculating the deformation between the observed object and its corresponding shape prior.
However, the static shape prior reconstruction faces difficulties in handling large intra-class variations.
Firstly, objects within the same category can still exhibit significant shape differences, as shown in Fig.\ref{fig:deformation}(A), which considerably increases the difficulty of template deformation and leads to inaccurate reconstruction.
Secondly, the 6-DoF (i.e., six degrees of freedom) rotation of the observed object in space, as illustrated in Fig.\ref{fig:deformation}(B), significantly impacts the extraction of template deformation features by introducing spatial rotation information interference in the real world.

In response to the above two challenges, numerous methods~\cite{chen2020learning,deng2021deformed,lin2022sar} have been proposed that consider deforming templates to better cope with the variability between instances, which can be divided into two groups: geometric shape deformation and spatial rotation deformation.
Geometric shape deformation approaches~\cite{chen2020learning,chen2021sgpa,zhang2022ssp} mainly focus on extracting geometric features from the observed object point cloud, adaptively deforming the prior template based on these features, and subsequently comparing the deformed template with the observed object point cloud to derive the 6D pose.
However, as validated in a few works~{\cite{park2019deepsdf,deng2021deformed}}, these methods usually adopt explicit encoder-decoder (e.g., PointNet-like) structures, which may fail to capture more complex shape deformation patterns,
limiting their generalization ability when addressing heterogeneous instances within the same category.
Moreover, the inferior reconstruction results from the shape template directly impair the pose estimation performance.

Spatial rotation deformation methods~\cite{deng2022icaps,lin2022sar} mainly involve rotating the template based on extracted spatial rotation features of the observed object point cloud and aligning the rotated template with the prior template for pose estimation.
However, since template deformation requires the observed object's spatial rotation features, pose estimation and template deformation are tightly coupled.
Thus, the accuracy of pose estimation and template deformation depends on the precise representation of the observed object's spatial rotation features, which can be relatively challenging in complex and diverse real-world scenarios, especially with excessive rotation angles and limited observable geometric structures.

In addition, most aforementioned and commonly used methods~\cite{tian2020shape,chen2020learning,di2022gpv,lin2022sar,zhang2022ssp} rely on separate segmentation modules to detect instance regions as input for subsequent pose and reconstruction modules.
This non-end-to-end inference approach isolates each object from the scene without important environmental contexts and has trouble handling problems in real-world environments, such as object occlusions and shape variations.

To tackle the above problems, we propose a novel category-level pose estimation and shape reconstruction method, {\dtf}.
To effectively capture the geometric deformation between the observed instance and the corresponding categorical template, we design a deformable template field, which introduces two implicit neural fields to represent the template features of the category and the continuous geometric deformation features of the observed instance, providing better handling of heterogeneous variation, as shown in Fig.\ref{fig:deformation}(A).
Our pose regression module, by extracting the 6D pose features of the instance point cloud and combining the aforementioned template and deformation features, can more accurately estimate the pose of each instance within the category.
In order to reduce the coupling between pose estimation and template deformation prediction, we train to effectively represent the rotation properties of objects through a designed shape-invariant training strategy, as shown in Fig.\ref{fig:deformation}(B).
Moreover, we utilize multi-view sampling training to further enhance the network's ability to handle varying poses in real-world scenarios.
Our entire model supports end-to-end inference, aided by a multi-modal representation extraction module for detecting instance keypoints, generating geometric feature maps, and predicting instance segmentation masks.
By utilizing this structure, the understanding of the entire scene is enhanced compared to only performing geometric analysis on regions of interest.
Our main contributions are summarized as follows:

\begin{itemize} [itemsep=1pt,topsep=1pt,parsep=1pt,leftmargin=10pt]
    \item We propose a novel network named {\dtf}, which reconstructs 3D shapes in fine detail and obtains accurate pose estimations.
    \item We design a deformable template field to represent the template prior and the corresponding deformation of the observed instance, which enables continuous deformation of arbitrary objects of the same category.
    \item We design a training strategy based on the shape-invariant training strategy, which can decouple between pose and deformation features to significantly improve pose estimation performance.
    \item Our method achieves state-of-the-art performance on the CAMERA25 and REAL275 benchmarks and is shown to improve grasping performance on a real robot arm.
\end{itemize}

\section{Related Work}
\paragraph{Category-level 6D pose estimation}
The category-level pose estimation aims to predict the 6D pose of the unseen object, which is challenging due to the lack of original CAD models of the observed objects.
As a seminal work in this research direction, Wang et al.~\cite{wang2019normalized} propose Normalized Object Coordinate Space (NOCS) that normalizes objects.
Some methods~\cite{chen2021fs,irshad2022centersnap} directly regress pose parameters. For example,
FS-Net~\cite{chen2021fs} regresses the pose using an orientation-aware autoencoder with 3D graph convolution;
CenterSnap~\cite{irshad2022centersnap} utilizes an autoencoder to encode the object shape and train a backbone to regress latent code, pose, and size.
These methods are sensitive to the point cloud input and have difficulty capturing features of continuous shape transformations.
A few methods~\cite{tian2020shape,wang2021category,chen2021sgpa} leverage category-level shape priors and learn the dense correspondence.
CR-Net~\cite{wang2021category} utilizes category prior to get category relation and designs a recurrent reconstruction network.
SGPA~\cite{chen2021sgpa} utilizes the shape prior and proposes a structure-guided prior adaption scheme, dynamically learning the prior according to observed objects. These constraints on categorical geometric features limit the generalization ability of the model partly.

\paragraph{Shape representation}
Point cloud~\cite{foley1996computer}, mesh~\cite{baumgart1975polyhedron}, and voxel~\cite{foley1996computer} are common shape representations in computer graphics.
Point clouds represent the shape of objects in a straightforward way, but they can not represent the accurate surface of the object~\cite{park2019deepsdf}.
Meshes can provide more accurate surface information, but one mesh can only represent a single object’s shape, which limits the scalability.
Voxels in 3D space are similar to pixels in 2D space, and some voxel-based efforts are transferred from pixel-based methods in 2D space, i.e., using 3D convolutional neural networks~\cite{wu20153d,choy20163d}.
Moreover, some methods use voxels to represent signed distance function~(SDF)~\cite{dai2017shape,stutz2018learning,chen2019learning}.
However, these methods require high memory, resulting in limited resolutions.
Recent works~\cite{park2019deepsdf,chen2019learning,mescheder2019occupancy} use binary implicit surface representations and learn implicit functions for object shapes, which predicts whether the 3D point is on the surface.
DeepSDF~\cite{park2019deepsdf} learns continuous SDF representation of shapes in a class, which can achieve smooth deformation between two different shapes in the same class by interpolation.
DIF-Net~\cite{deng2021deformed} proposed deformation fields, which generate 3D shapes with high fidelity and establish dense correspondences among different shapes.
Inspired by the above methods, we design a novel learnable implicit neural field, which can represent the continuous transformation of shape and pose between templates and instances, thereby improving the performance of pose estimation and reconstruction.

\section{Method}
\begin{figure*}[t]
  \centering
  \includegraphics[width=0.91\linewidth]{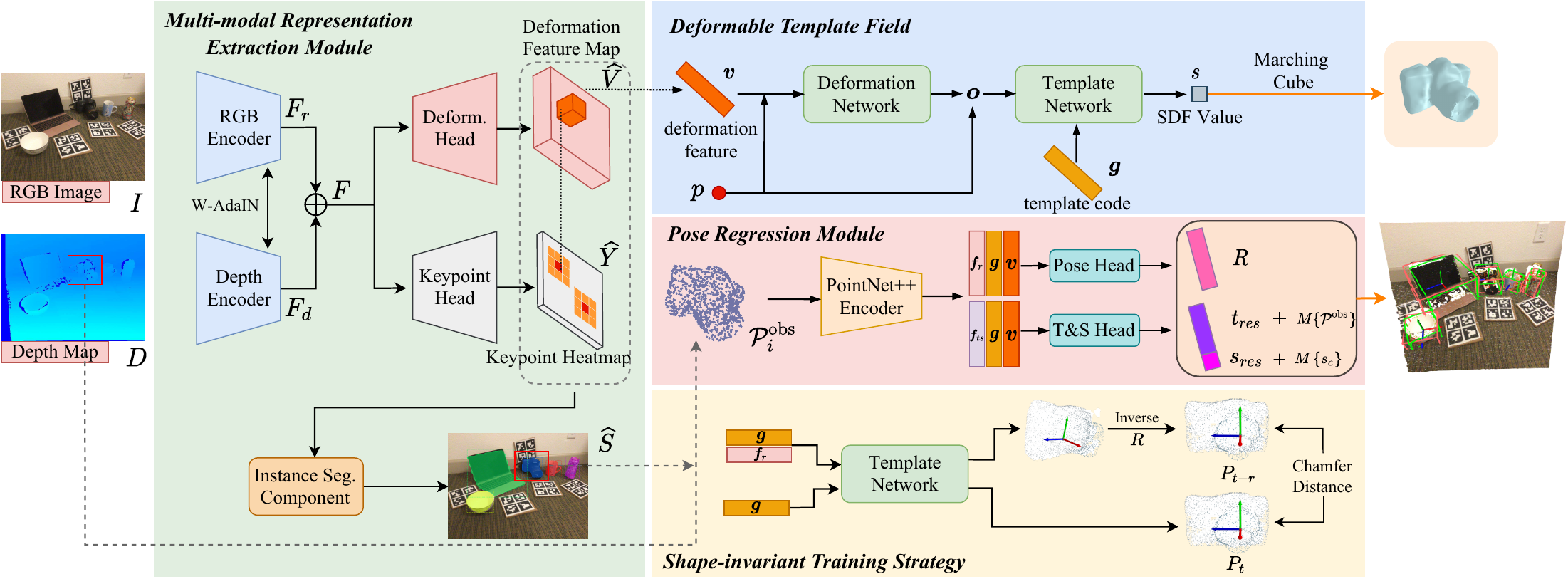}
  \caption{Architecture Overview of proposed DTF-Net.
    The architecture consists of three main components: the  multi-modal representation extraction module, the deformable template field, and the pose regression module.
    The multi-modal representation extraction module~(Section~\ref{sec:OFEM}) detects the keypoints of the instances.
    The deformable template field~(Section~\ref{sec:dtf}) predicts the spatial deformation of the observed objects to the template implicit field.
    The pose regression module~(Section~\ref{sec:prm}) inputted the point cloud and combined deformable template features to estimate the 6D pose and size of the instances in the scene.
    The shape-invariant training strategy~(Section~\ref{sec:training}) can further decouple the geometric features and rotation features.}
  \label{fig:architecture}
\end{figure*}

In this section, we present the proposed {\dtf}.
Its structure is presented in Fig.\ref{fig:architecture}. {\dtf} adopts an RGB image $I\in \mathbb{R}^{W \times H \times 3}$ and a corresponding depth map  $D\in \mathbb{R}^{W \times H \times 1}$ to infer the 3D rotation, 3D translation, and size $\{\boldsymbol{R}, \boldsymbol{t}, s\}$ of multiple objects in the scene and completes the 3D shape reconstruction of each object.

{\dtf} consists of three components:

\begin{itemize} [itemsep=1pt,topsep=1pt,parsep=1pt,leftmargin=10pt]
    \item \textit{Multi-modal Representation Extraction Module} (Section~\ref{sec:OFEM}) acquires a keypoint heatmap $\widehat{Y} \in[0,1]^{\frac{W}{R} \times \frac{H}{R} \times N}$, a deformation feature map $\widehat{V}\in \mathbb{R} ^{\frac{W}{R} \times \frac{H}{R} \times 256}$, and an instance segmentation map $\widehat{S} \in \mathbb{R}^{{W} \times {H} \times 1}$ using a pair of RGB $I$ and depth $D$ images.
    \item \textit{Deformable Template Field} (Section~\ref{sec:dtf}) consists of a deformation network~(\textit{D}) and a template network~(\textit{T}).
    The template network learns the template feature $\boldsymbol{g}$ for each category from the 3D model collection $\{\mathbf{M}\}$ of the respective categories, while the deformation network captures the shape deformation feature  $\boldsymbol{v}$ of each instance relative to the template.
    \item \textit{Pose Regression Module} (Section~\ref{sec:prm}) processes the depth map of the instance segmentation mask region as the point clouds ${\mathcal{P}^{obs}}$ of target objects, and predicts rotation, position, and size information $\{\boldsymbol{R}, \boldsymbol{t}, s\}$ by learning its latent features in conjunction with shape deformation features and template features.
\end{itemize}

\subsection{Multi-Modal Representation Extraction}
\label{sec:OFEM}
We design a multi-modal representation extraction module to obtain the keypoints heatmap, geometric deformation feature map, and segmentation masks in the whole scene, and at the same time as the input of subsequent modules for  learning estimation.
This allows the subsequent pose estimation module to effectively capture the global information of the scene as well.
It includes a keypoint extraction component and an instance segmentation component.

\para{Keypoint extraction component}
We adopt an anchor-free network~\cite{zhou2019objects, irshad2022centersnap} architecture to extract the required features from the scene.
Given a set of RGB data $I$ and depth data $D$ as input, the encoder performs downsampling to generate RGB and depth features ${F_{r},F_{d}} \in \mathbb{R}^{\frac{W}{R} \times \frac{H}{R} \times C}$, where $R=8$ and $C=128$ represent the downsampling ratio and the channel number, respectively.
In order to more effectively fuse the RGB and depth data originating from two distinct distributions, the W-AdaIN structure~\cite{wang2022rgb} is incorporated into each layer of the encoder.
We adopt channel connection to fuse $F_{r}$ and $F_{d}$, to obtain a new feature $F\in \mathbb{R}^{\frac{W}{R} \times \frac{W}{R} \times 2C}$.
And we feed $F$ into the keypoint head and deformation head to obtain the keypoint heatmap $\widehat{Y} \in[0,1]^{\frac{W}{R} \times \frac{H}{R} \times N}$ and geometric deformation feature map $\widehat{V}\in \mathbb{R} ^{\frac{W}{R} \times \frac{H}{R} \times 256}$.
Here, $N$ is the category number, and  $\widehat{Y}_{w,h,n} ={0/1}$ is background or keypoint.

Subsequently, we select 64 peaks whose value is greater or equal to its 8-connected neighbors of the $\widehat{Y} \in[0,1]^{\frac{W}{R} \times \frac{H}{R} \times N}$ as center points of object $\hat{\mathcal{P}}=\left\{\left(\hat{x}_i, \hat{y}_i\right)\right\}_{i=1}^{64}$.
And we can obtain the corresponding deformation features $\hat{\boldsymbol{v}}_{\left(\hat{x}_i, \hat{y}_i\right)}\in \mathbb{R}^{256}$ at those locations, which serve as inputs for subsequent processing.

\para{Instance segmentation component}
We directly generate instance segmentation mask $\widehat{S} \in \mathbb{R}^{{W} \times {H} \times 1}$ centered around the keypoint heatmap and geometric deformation feature map in the previous step.
We upsample the keypoint heatmap and geometric deformation feature map from $(W/R, H/R)$ back to $(W, H)$, generating ${S}_g \in \mathbb{R}^{{W} \times {H} \times 1}$ and ${S}_l \in \mathbb{R}^{{W} \times {H} \times 1}$, respectively.
The ${S}_g$ is generated from the keypoint heatmap and offers a global prediction of the scene foreground mask distribution through the heatmap representation of instance locations.
The ${S}_l$ is generated from the geometric deformation feature map and can better capture object geometric details.

By applying a sigmoid function, we normalize the values of ${S}_g $ and ${S}_l $ to the range of (0, 1) and fuse them by the Hadamard product, obtaining the final segmentation result $\widehat{S}$. We then optimize this result using an independent semantic segmentation loss function.

\subsection{Deformable Template Field}
\label{sec:dtf}

\begin{figure}[t]
  \centering
  \includegraphics[width=0.9\linewidth]{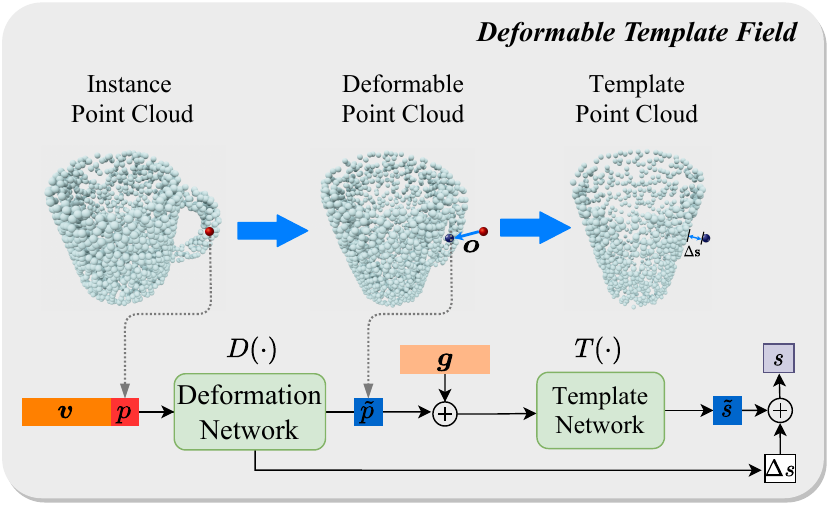}
  \caption{The architecture of the deformable template field.
  }
  \label{fig:tdf}
\end{figure}

Our goal is to learn category-level geometric template code $\boldsymbol{g}$ and capture the geometric deformation feature $\boldsymbol{v}$ between the observed instances and their corresponding templates through training from the available 3D object models $\{\mathbf{M}\}$.
Based on the SDF representation used in Implicit Neural Fields~\cite{zheng2021deep, deng2021deformed}, we propose the deformable template field to generate categorial prior templates and establish spatially continuous variation with the observed objects.
As shown in Fig.\ref{fig:tdf}, the deformable template field is a decoder composed of the deformation network~(\textit{D}) and template network~(\textit{T}).

\para{Deformation network}
We employ this module to obtain the geometric deformation features to represent the continuous shape variation space between observed instances and prior templates:
\begin{equation}
    D(\mathbf{p}, \boldsymbol{v}_j)=(\mathbf{o}, \Delta {s}): \mathbf{p} \in \mathbb{R}^3, s \in \mathbb{R},
\label{eq:deformation}
\end{equation}
where $\mathbf{o}$ is the translation offset for a point $\mathbf{p}$  in space and $\Delta {s}$ is the distance correction value for some heterogeneous objects in the same category.
Through the backpropagation of Eq.~\eqref{eq:deformation}, we can obtain the deformation feature $\boldsymbol{v}_j$ of each instance relative to its corresponding categorial template.

\para{Template network}
To represent the general template shape features for multiple categories, we improve DeepSDF~\cite{zheng2021deep} to map a latent code $\boldsymbol{z}_{j}^{i}$ and a 3D point $\mathbf{p}$ to a scalar ${s}$:
\begin{equation}
    T(\mathbf{p}, \boldsymbol{z}_{j}^{i})= {s}: \mathbf{p} \in \mathbb{R}^3, {s} \in \mathbb{R},
\label{eq:template}
\end{equation}
where ${s}$ is the closest distance between point $\mathbf{p}$ and the instance object surface of $\mathbf{M}$.
$T(\cdot)$ is a continuous representation, and $T(\cdot) = 0$ implicitly denotes the object's shape surface.
Given an obtainable object model set $\left\{\mathbf{M}_j^i\right\}_N^C $, where $\mathbf{M}_j^i$ denotes the 3D point cloud model of instance $j$ from category $i$.
The latent code $\boldsymbol{z}_j^i$ can represent the geometric feature for each model by back-propagation of Eq.~\eqref{eq:template}.
Then we can calculate the template code $ \boldsymbol{g}^{i}$ for each category:
\begin{equation}
    \boldsymbol{g}^{i}=\dfrac{1}{N}\sum ^{N}_{j=1} \boldsymbol{z}_{j}^{i}( i= 1,2,\ldots,C).
\label{eq:g}
\end{equation}

By shared weights of different objects and the constraints of categorical latent template codes, the template network can represent geometric information for different categories.

Overall, the deformable template field uses the deformation net to obtain the geometric deformation of the instance and then predict the SDF value of the template via the template network and categorial template code:
\begin{equation}
    T\left(D(\boldsymbol{v}_j^i, \mathbf{p}),\boldsymbol{g}^{i}\right)=T\left(\mathbf{p}+(\mathbf{o}, \Delta {s}), \boldsymbol{g}^{i}\right) = s .
\end{equation}

This procedure creates a continuous mapping between an object instance and its corresponding template, thereby obtaining the geometric shape of the instance.

\subsection{Pose Regression}
\label{sec:prm}
We use the segmentation masks to separate the instance objects from the depth map and then use the camera intrinsics to convert the instance objects into point clouds $\{\mathcal{P}^{obs}_{i}\}_{i=1}^{k}$, as shown in gray dotted lines of Fig.\ref{fig:architecture}.
Here, $k$ is the number of instances in the current scene.
$\mathcal{P}^{obs}_{i}$ obtains the spatial rotation features $\boldsymbol{f}_r$ of rotation and $\boldsymbol{f}_{ts}$ of size and translation by encoder of the PointNet++ structure~\cite{qi2017pointnet++}.
We feed $\boldsymbol{f}_r$ combined template code $\boldsymbol{g}$ and deformation feature $\boldsymbol{v}$ to pose head and predict the 9-D vector, which represents 6D pose rotation $\hat{\boldsymbol{R}} \in SO(3)$ for the object.
Also, we estimate the size and translation residuals $t_{res}$ and $s_{res}$ via size\&translation head with $\boldsymbol{f}_{ts}$ and $[\boldsymbol{g}, \boldsymbol{v}]$.
The final predictions of size and translation are $\hat{s} = s_{res} + M\left\{s_c\right\}$ and $\hat{t} = t_{res} + M\left\{\mathcal{P}^{obs}_{i}\right\}$, where $M\left\{s_c\right\}$ is the mean size of the prior category, and $M\left\{\mathcal{P}^{obs}_{i}\right\}$ is the mean observed point cloud of the object.
During the training process,
we combine the template net to decouple the $\boldsymbol{f}_r$ from the object's geometric features as much as possible in order to ensure the accuracy of the rotation estimation, which will be described in detail in Section~\ref{sec:training}.

\subsection{Efficient and Integrated Training Strategy}
\label{sec:training}
\para{Overall pipeline of training}
We first train the template network to obtain a latent template code for each category and then train the whole deformable template field, which can establish a continuous shape variation between a specific object and the template, which loss function can be formulated as:
\begin{equation}
    \mathcal{L}_{tdf} =\sum_i\sum_j\sum_{\mathbf{p}_c \in \Omega}\left|T\left(D(\boldsymbol{v}_j^i, \mathbf{p}_c), \boldsymbol{g}^i\right)-\bar{s}\right| + \sum_i\sum_j{\left\|\boldsymbol{v}_j^i\right\|}_2^2,
\end{equation}
where $\bar{s}$ is the ground truth of SDF value, and $\Omega$ is the 3D space.  The first part of the equation computes the error between the predicted shape and the actual shape, while the second part constrains the complexity of the shape deformation features to enhance the model's generalization ability. In the training phase, all point clouds $\mathbf{p}$ of object models should be converted to the canonical coordinate:
\begin{equation}
    \mathbf{p}_c = \boldsymbol{R}^T(\mathbf{p} -\mathbf{t})/L,
\end{equation}
where $\boldsymbol{R}$ and $\mathbf{t}$ are the ground truth of the rotation and translation, and $ L $ is the diagonal length. This transformation streamlines the training process by enabling the model to learn object features invariant to size, orientation, and position.

For keypoint extraction branch, $\widehat{Y}$ is spread by the keypoints of the target object onto the heatmap through the Gaussian kernel and trained with the focal loss $\mathcal{L}_{k}$~\cite{zhou2019objects}. The loss function is as follows:
\begin{equation}
\mathcal{L}_{\text {k}} = -\frac{1}{M} \sum_{w h n}\left\{\begin{array}{ll}
{(1-\widehat{Y}_{w h n})}^\alpha \log (\widehat{Y}_{w h n}) & \text {if } Y_{w h n}=1, \\
{(1-\widehat{Y}_{w h n})}^\beta \widehat{Y}_{w h n}^\alpha \log (1-\widehat{Y}_{w h n}) & \text {otherwise,}
\end{array}\right.
\end{equation}
where $M$ is the number of keypoints, $\widehat{Y}_{whn}$ is the score for category $n$ at location $(w, h)$ in the predicted heatmap, and $\alpha$ and $\beta$ are the hyper-parameters of the focal loss.

Similarly, we splat the geometric deformation feature $\boldsymbol{v}$ obtained by deformable template fields through Gaussian kernel to generate the feature map ${V}$ to optimize the geometric deformation map $\widehat{V}$ predicted by keypoint extraction branch. The loss function is:
\begin{equation}
    \mathcal{L}_{v}={\|\widehat{V}-V\|}_2^2.
\end{equation}

For instance segmentation branch, the loss function is:
\begin{equation}
\mathcal{L}_{\text {seg}} =
\frac{1}{N} \sum_{k=1}^N \left[-\left(\widehat{S} \log({S} ) + (1 - \widehat{S}) \log(1 - {S})\right)\right],
\end{equation}
where $N$ is the number of objects,  ${S}$ and $\widehat{S}$ are ground truth and predict results of the segmentation mask, respectively.

We use Chamfer distance to measure the dissimilarity between the point clouds transformed using the predicted pose and the ground truth pose to define the loss function for pose prediction:
\begin{equation}
\mathcal{L}_{\text {pose }}=\frac{1}{M} \sum_i \min _{0<j<M}\left\|s \left(R x_i+t\right)-\hat{s} \left(\hat{R} x_j+\hat{t}\right)\right\|,
\end{equation}
where $x_i \in \mathcal{R}^3 $ is a 3D point of the object, $s$ and $\hat{s}$ are the sizes, $R, \hat{R} \in \mathcal {SO}(3)$,  $t, \hat{t}  \in \mathcal{R}^3 $, and $M$ is the number of sampled points.

\begin{figure}[!t]
  \centering
  \includegraphics[width=0.95\linewidth]{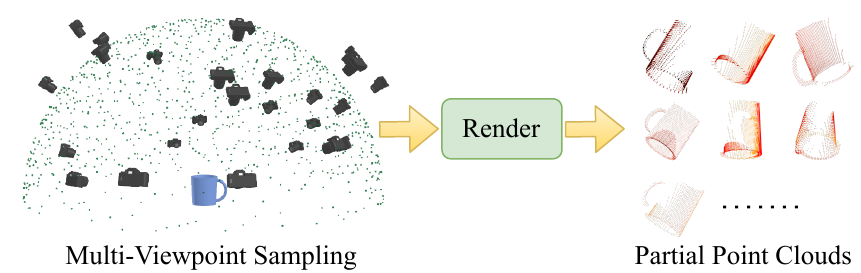}
  \caption{Viewpoint sampling process. The viewpoints are sampled from a semisphere centered on the instance model.}
  \label{fig:viewpoint}
\end{figure}

\para{Shape-invariant training strategy}
The encoder represents the observed point cloud as a spatial rotation embedding $\boldsymbol{f_{r}}$, which is then combined with the object's latent template code $\boldsymbol{g}$ from the deformable template field. Theoretically, since $\boldsymbol{f_{r}}$ encapsulates spatial rotation information, it should only influence the object's rotation, not its shape. Consequently, the template network's output, which merges $\boldsymbol{g}$ and $\boldsymbol{f_{r}}$, should only adjust the template's spatial rotation without affecting its geometry.

Based on the aforementioned analysis, in the training phase shown in Fig.\ref{fig:architecture},
we add the generated rotation features $\boldsymbol{f_{r}}$ to the template code $\boldsymbol{g}$, yielding the rotated template via the template network.
This rotated template is then subjected to an inverse rotation operation to obtain an inverse-rotated template using the ground truth rotation label, aligning it back to the canonical coordinates.
Two point clouds with 1024 points on the object surface, $P_t$ and $P_{t-r}$, are sampled on the template and inverse-rotated template, respectively.
To make these two point clouds as shape congruent as possible, the Chamfer distance is employed between the template point clouds $P_t$ and inverse-rotated point clouds $P_{t-r}$:
\begin{equation}
\begin{aligned}
    \mathcal{L}_{t-r} & = \frac{1}{P_t} \sum_{x \in P_t} \min _{y \in P_{t-r}}{\|x-y\|}_2^2 + \frac{1}{P_{t-r}} \sum_{y \in P_{t-r}} \min _{x \in P_t}{\|y-x\|}_2^2.
\end{aligned}
\end{equation}

\para{Viewpoint sampling}
We design a sampling method based on viewpoint transformation to emulate potential real-world scenarios, as shown in Fig.\ref{fig:viewpoint}.
We place the object horizontally on the table and uniformly sample $N$ camera viewpoints $\left\{\boldsymbol{C}_i\right\}_{i=1}^N$ from a semisphere centered on the object with the radius $\mathbf{r} $.
PyTorch3D~\cite{johnson2020accelerating} is used to render the partial point cloud.
It is sufficient to have the object's 6D pose $\mathbf{P}_0$ from a single viewpoint $\boldsymbol{C}_0$, as the poses in the other scenes can be deduced via camera pose RT transformations:
\begin{equation}
    \mathbf{P}_i=\boldsymbol{C}_i^{-1} \boldsymbol{C}_0 \mathbf{P}_0,
\end{equation}
where $\mathbf{P}_i$ is the object pose at viewpoint $i$ and $\boldsymbol{C}_i$ is the camera pose.

\para{Overall loss}
The overall loss function $\mathcal{L}$ for training is:
\begin{equation}
    \mathcal{L}=\lambda_1 \mathcal{L}_{\text {pose }}+\lambda_2 \mathcal{L}_{t-r}+\lambda_3 \mathcal{L}_{\text {v }}+\lambda_4 \mathcal{L}_{\text {seg}}+ \lambda_5 \mathcal{L}_{\text {k}}.
\end{equation}

\begin{figure*}[!b]
  \centering
  \includegraphics[width=0.92\linewidth]{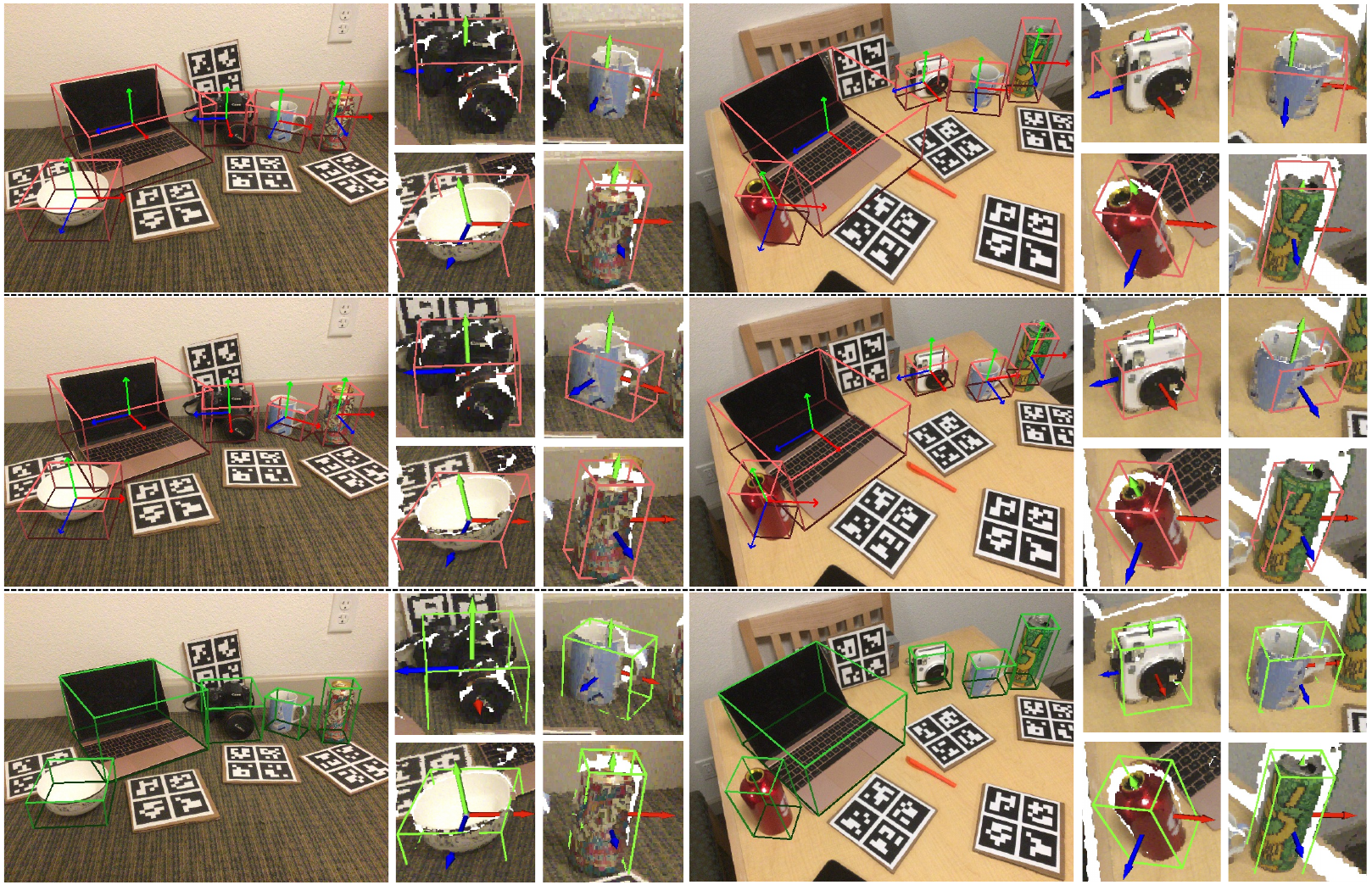}
  \caption{Pose and size estimation results of CenterSnap~\cite{irshad2022centersnap} (top), our proposed method (middle), and ground truths (bottom) on REAL275. The green and red boxes are the ground truths and the predictions, respectively.}
  \label{fig:sota-fig}
\end{figure*}

\begin{table*}[!t]
\caption{
  Comparison with state-of-the-art methods on REAL275 and CAMERA25.
  `P.S.' denotes that the pre-processing via segmentation task is required.
   The speed metric refers exclusively to the 6D pose estimation process.
}
\label{tab:sota}
\centering
\resizebox{0.85\textwidth}{!}{
\begin{tabular}{c|c|c|ccc|cccc|c}
    \toprule
    Dataset & Method &  P.S. & $IoU_{25}$ & $IoU_{50}$ & $IoU_{75}$ &  $5^{\circ}\ 2cm$ & $5^{\circ}\ 5cm$ & $10^{\circ}\ 5cm$ & $10^{\circ}\ 10cm$ & Speed (FPS) \\
    \midrule
    \multirow{10}{*}{REAL275}
    & NOCS~\cite{wang2019normalized} &  & $\underline{84.9}$ & $80.5$ & $-$ & $7.2$ & $10.0$ & $25.2$ & $26.7$ & $5$ \\
    & ShapePrior~\cite{tian2020shape} & \checkmark & $83.4$ & $77.3$ & $53.2$ & $19.3$ & $21.4$ & $54.1$ & $-$ & $4$ \\
    & CASS~\cite{chen2020learning} & \checkmark & $84.2$ & $77.7$ & $-$ & $-$ & $23.5$ & $58.0$ & $58.3$ & $-$ \\
    & DualPoseNet~\cite{lin2021dualposenet} & \checkmark & $-$ & $79.8$ & $62.2$ & $29.3$ & $35.9$ & $66.8$ & $-$ & $2$ \\
    & SGPA~\cite{chen2021sgpa} & \checkmark & $-$ & $80.1$ &&  $\underline{35.9}$ & $39.6$ & $70.7$ & $-$ & $3$ \\
    & CenterSnap~\cite{irshad2022centersnap} &  & $83.5$ & $-$ & $-$ & $-$ & $27.2$ & $58.8$ & $64.4$ & $\mathbf{40}$ \\
    & GPV-Net~\cite{di2022gpv} & \checkmark & $84.2$ & $\underline{83.0}$  & $\underline{64.4}$ & $32.0$ & $\underline{42.9}$ & $73.3$ & $74.6$ & $20$ \\
    & SAR-Net~\cite{lin2022sar} & \checkmark & $-$ & $79.3$ & $62.4$ & $31.6$ & $42.3$ & $68.3$ & $-$ & $-$ \\
    \cmidrule(lr){2-11}
    & {\dtf}~(Ours) &  & $83.7$ & $80.6$ & $63.8$ & $30.7$ & $41.8$ & $\underline{75.8}$ & $\underline{77.4}$ & $\underline{24}$ \\
    & {\dtf}~(IR) &  & $\mathbf{86.1}$ & $\mathbf{84.0}$ & $\mathbf{66.9}$ & $\mathbf{39.3}$ & $\mathbf{52.1}$ & $\mathbf{79.7}$ & $\mathbf{81.0}$ & $3$ \\
    \cmidrule{1-11}
    \multirow{7}{*}{CAMERA25}
    & NOCS~\cite{wang2019normalized} &  & $91.4$ & $85.3$ & $-$ & $-$ & $38.8$ & $61.7$ & $62.2$ & $5$ \\
    & ShapePrior~\cite{tian2020shape} & \checkmark & $-$ & $93.2$ & $83.1$ & $54.3$ & $59.0$ & $81.5$ & $-$ & $4$ \\
    & DualPoseNet~\cite{lin2021dualposenet} & \checkmark & $-$ & $92.4$  & $86.4$ & $64.7$ & $70.7$ & $84.7$ & $-$ & $2$ \\
      & SGPA~\cite{chen2021sgpa} & \checkmark & $-$ & $93.2$  & $\mathbf{88.1}$ & $70.7$ & $74.5$ & $\mathbf{88.4}$ & $-$ & $3$ \\
    & CenterSnap~\cite{irshad2022centersnap} &  & ${93.2}$ & $92.3$ & $-$ & $-$ & $63.0$ & $79.5$ & $87.9$ & $\mathbf{40}$ \\
    & SAR-Net~\cite{lin2022sar} & \checkmark & $-$ & $86.8$ & $79.0$ & $66.7$ & $70.9$ & $80.3$ & $-$ & $-$ \\
    \cmidrule(lr){2-11}
    & {\dtf}~(Ours) & & $\mathbf{95.1}$ & $\mathbf{94.5}$ & $87.5$ & $\mathbf{71.2}$ & $\mathbf{77.9}$ & ${85.5}$ & $\mathbf{89.6}$ & ${24}$   \\
    \bottomrule
\end{tabular}
}
\end{table*}

\section{Experiments}

\subsection{Experiment Settings}
\para{Datasets}
We evaluated our method on two datasets, which are a synthetic dataset (CAMERA25) and a real dataset (REAL275)~\cite{wang2019normalized}.
CAMERA25 contains 300K composite images, where 25K are used for evaluation.
There are 1085 object instances and 184 different instances from 6 categories for training and evaluation, respectively, including \textit{bottle}, \textit{bowl}, \textit{camera}, \textit{can}, \textit{laptop}, and \textit{mug}.
REAL275 contains 4300 real-world images of 7 scenes for training and 2750 real-world images of 6 scenes for evaluation.

\para{Metrics}
We computed the average precision of 3D Intersection-Over-Union~(IoU) at different thresholds of 25\%, 50\%, and 75\% for object detection.
The average precision of 6D pose is evaluated at $\mathrm{x}^{\circ}\ \mathrm{y}cm$, including $\mathrm{5}^{\circ}\ \mathrm{2}cm$, $\mathrm{5}^{\circ}\ \mathrm{5}cm$, $\mathrm{10}^{\circ}\ \mathrm{5}cm$, and $\mathrm{10}^{\circ}\ \mathrm{10}cm$, to measure errors in rotation and translation.
Meanwhile, we followed common practices~\cite{tian2020shape} to evaluate the shape reconstruction.

\para{Implementation details}
For the multi-modal representation
extraction module, we applyed DLA-34~\cite{yu2018deep} as the backbone. The RGB-D images were reshaped $512 \times 512$ as input resolution. The down-sampling stride $R = 4 $.
We uniformly sampled each 3D model from the CAMERA25 training set into 2048 points to train the deformable template field.
The overall architecture was trained for 100 epochs with a batch size of 32.
We employed random uniform noise, random scaling, random rotation, and translation for point clouds.
We set the hyper-parameter $\left\{\lambda_1, \lambda_2,  \lambda_3, \lambda_4, \lambda_5\right\}=\{10, 2, 5, 1, 1\}$.

\subsection{Comparisons with State-of-the-Art Methods}
\label{sec:exp-sota}
We compared the proposed {\dtf} method with several state-of-the-art methods.
The quantitative comparison results are reported in Tab.\ref{tab:sota}, and the qualitative comparisons are shown in Fig.\ref{fig:sota-fig}.

The results presented in Tab.\ref{tab:sota} illustrate that our proposed method achieves state-of-the-art performance on both datasets while maintaining high inference efficiency. 
Notably, our method yields a precision of 75.8\% and 77.4\% on the $10^{\circ}\ 5cm$ and $10^{\circ}\ 10cm$ criteria, respectively, outperforming alternative methods substantially.
Meanwhile, we introduced a refinement strategy via iterative optimization and deformable net training between the object pose and the corresponding shape, as delineated in the DTF-Net~(IR) row of Tab.\ref{tab:sota}. 
Specifically, we aligned the reconstructed object point cloud with the predicted  pose point cloud and updated the pose. Then, we re-estimated the pose of the updated partial point cloud using the pose regression module and iterated the above process. 
This strategy can significantly boost the performance of our model.
In Fig.\ref{fig:sota-fig}, we compared our method with the pose estimation approach~\cite{irshad2022centersnap}, demonstrating our model's superior performance in detecting all instances and creating accurate bounding boxes with rotations.

\subsection{3D Shape Reconstructions}
\label{sec:exp-recon}
\begin{figure}[t]
  \centering
  \includegraphics[width=0.9\linewidth]{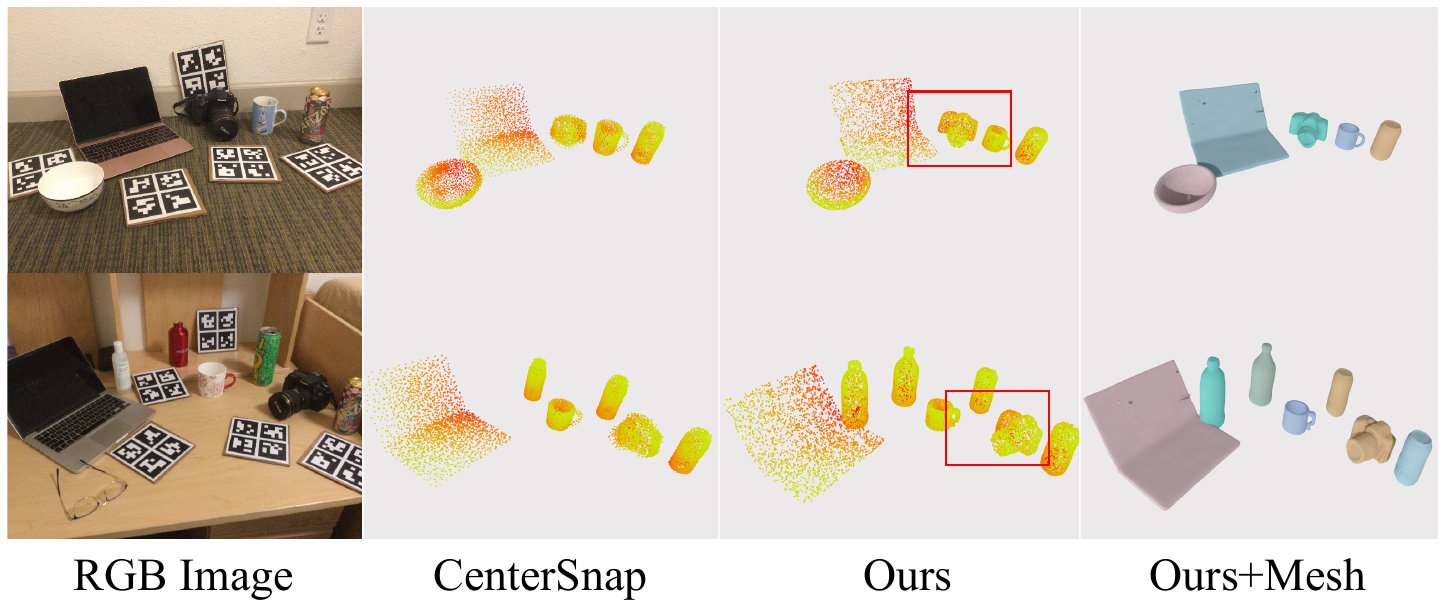}
  \caption{Visualizations of 3D shape reconstructions in a real scene from REAL275.}
  \label{fig:recon}
\end{figure}

Quantitative results of 3D point cloud reconstruction are tabulated in Tab.\ref{tab:recon}.
We compared our method with CASS~\cite{chen2020learning}, ShapePrior~\cite{tian2020shape}, and CenterSnap~\cite{irshad2022centersnap} by computing the Chamfer distance between the reconstructed point cloud and the ground truth point cloud.
To make settings consistent, we employed the deformable template field and marching cube~\cite{lorensen1987marching} to reconstruct the object mesh model and then uniformly sample 1024 points of the surface.
Owing to the precise shape reconstruction enabled by deformable template fields, our method outshines state-of-the-art techniques.
Fig.\ref{fig:recon} illustrates that our approach is capable of reconstructing smooth surface models for all objects within complex scenes.

\begin{table}[t]
\caption{Evaluation of point cloud reconstruction with the Chamfer distance~(CD).
The mean $\pm$ standard deviation results on six categories are shown. Unit: $1 \times 10^{-3}$.}
\label{tab:recon}
\centering
\resizebox{1\columnwidth}{!}{
  \begin{tabular}{cccc|c}
      \toprule
      CASS~\cite{chen2020learning}  & ShapePrior~\cite{tian2020shape} & CenterSnap~\cite{irshad2022centersnap} & FS-Net~\cite{chen2021fs} & {\dtf} \\
      \midrule
       $1.06\pm1.32$  & $3.17\pm 2.96$ & $1.50\pm 1.49$ & $0.86\pm 0.74$ & $\mathbf{0.71}\pm \mathbf{0.39}$\\
       \bottomrule
  \end{tabular}
}
\end{table}

\begin{figure}[t]
  \centering
  \includegraphics[width=1\linewidth]{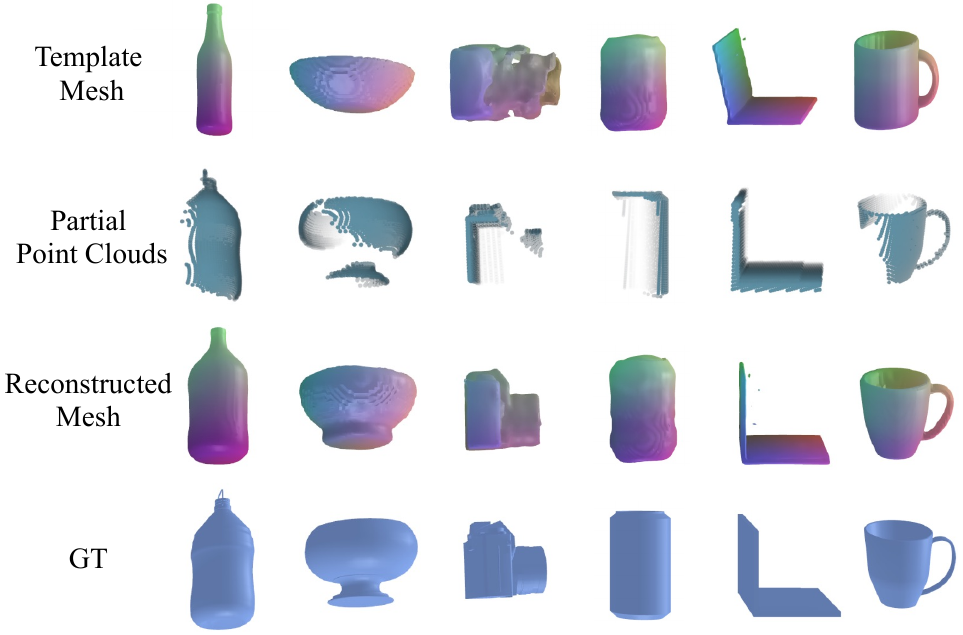}
  \caption{Visualizations of 3D shape reconstructions in mesh. }
  \label{fig:recon-dtf}
\end{figure}

Fig.\ref{fig:recon-dtf} showcases the templates obtained for six object categories using the Template Network and the deformed models generated by the Deformation Network from the partial point clouds of the scenes in the REAL275 dataset. The results clearly display that our approach manages to generate effective deformation outcomes even for heterogeneous objects that deviate from the templates, thereby facilitating fairly accurate 3D reconstructions.

\begin{figure}[!t]
  \centering
  \includegraphics[width=0.84\linewidth]{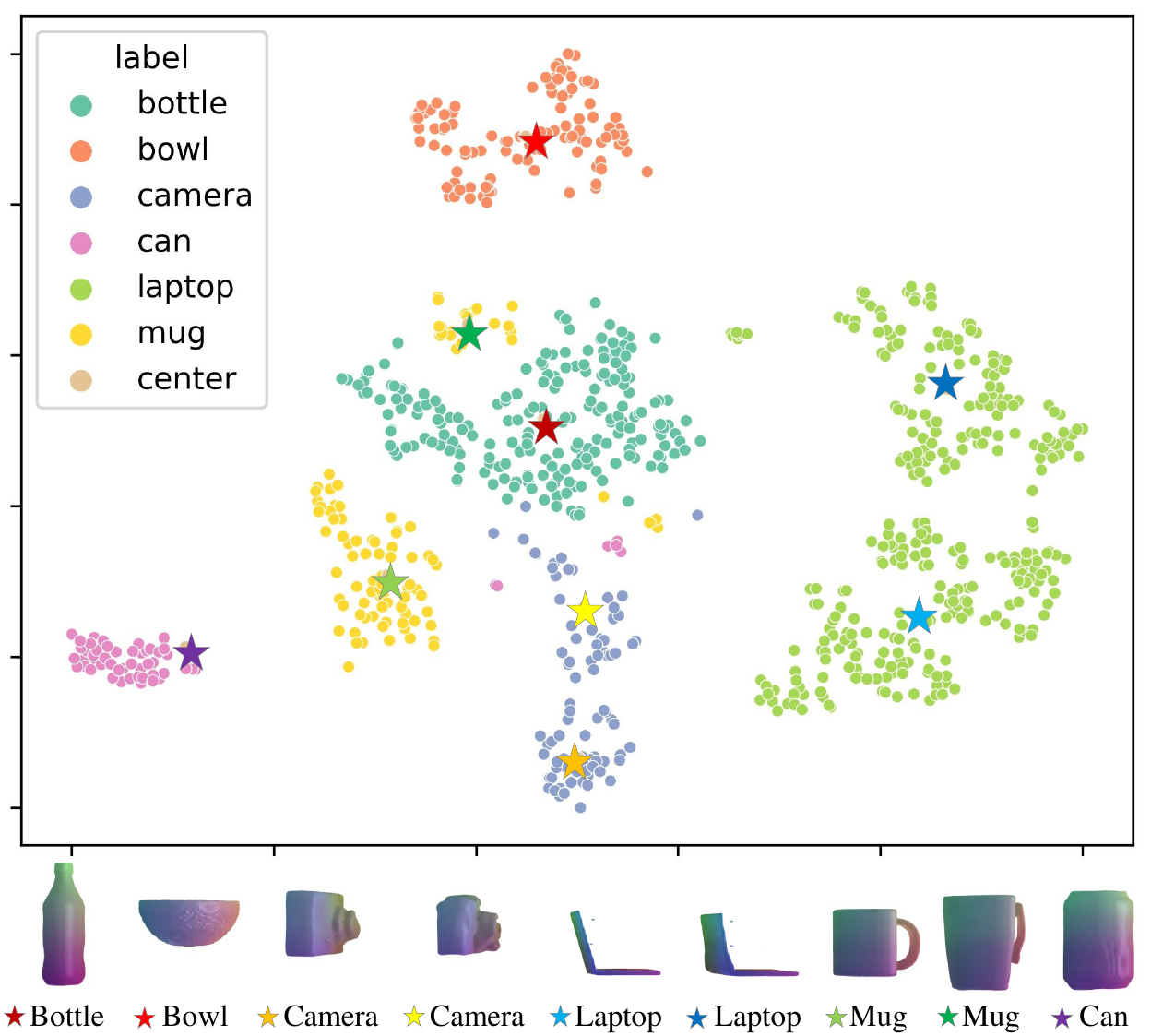}
  \caption{
    Geometric deformation feature visualization by t-SNE in the REAL275 test set (top) and
    reconstruction results of the central feature vectors of the nine clusters (bottom).
  }
  \label{fig:sne}
\end{figure}

We employed t-SNE to visualize the geometric deformation features ${\boldsymbol{v}}$ generated by {\dtf} in Fig.\ref{fig:sne}, demonstrating its effectiveness in representing geometric semantics across categories.
{\dtf} generates similar features for objects within the same category while simultaneously distinguishing some intra-class shape variations, such as the angle of laptop screens and types of cameras.

\subsection{Ablation Studies}
\label{sec:exp-abla}
\paragraph{Effect of the deformable template field}
We quantitatively evaluate the effectiveness of the deformable template field in Cases \#1-\#5 in Tab.\ref{tab:abla}.
We estimate the pose and size by the embeddings of the keypoints prediction module and reconstruct the object point clouds by an autoencoder for Case \#1.
Compared to the existing SDF~\cite{park2019deepsdf} (Case \#2), in Case \#3, our proposed deformation network,
combined with fixed template code, greatly improved the reconstruction performance in terms of the Chamfer distance (CD).
The combination with the template network further improved the performance of reconstruction and pose estimation, as shown in Cases \#4 and \#5.

\paragraph{Effect of the pose regression module}
Cases \#5-\#9 in Tab.\ref{tab:abla} show the efficacy of each part in the proposed regression module.
From the results, we can conclude
a) the segmentation module and PointNet++ encoder in Case \#6 can more effectively extract the pose information,
b) the spatial rotation features of the observed objects via the shape-invariant training strategy substantially improve the model performance, as shown in Case \#7,
c) the prediction of residual size and translation in Case \#8 enables higher accuracy than direct regression, and
d) viewsampling method can further improve the accuracy of pose estimation, as shown in Case \#9.

\begin{table}[!t]
\centering
\caption{Ablation study results of {\dtf} on REAL275.}
\label{tab:abla}
\resizebox{.85\columnwidth}{!}{
    \begin{tabular}{c|l|c|c}
        \toprule
        Case \# & Component & $IoU_{50}$ & ${CD}$ \\
        \midrule
        1 & Baseline & $69.3$  & $5.06$ \\
        \cmidrule(lr){1-4}
        2 & \#1 + DeepSDF & $71.7$  & $2.83$ \\
        3 & \#1 + Deformation Network & $71.3$  & $1.54$ \\
        4 & \#1 + Template Network & $74.0$  & $1.22$ \\
        5 & \#4 + Deformation Network & $73.9$  & $0.43$\\
        \cmidrule(lr){1-4}
        6 & \#5 + PointNet++ & $77.2$  & $-$\\
        7 & \#6 + Shape-invariant & $79.3$  & $-$\\
        8 & \#7 + Residual & $80.1$  & $-$\\
        9 & \#8 + ViewSampling & $80.6$  & $-$\\
        \bottomrule
    \end{tabular}
}
\end{table}

\subsection{Grasping Task in the Real World}
\label{sec:exp-grasp}
\begin{figure}[t!]
  \centering
    \includegraphics[width=0.95\linewidth]{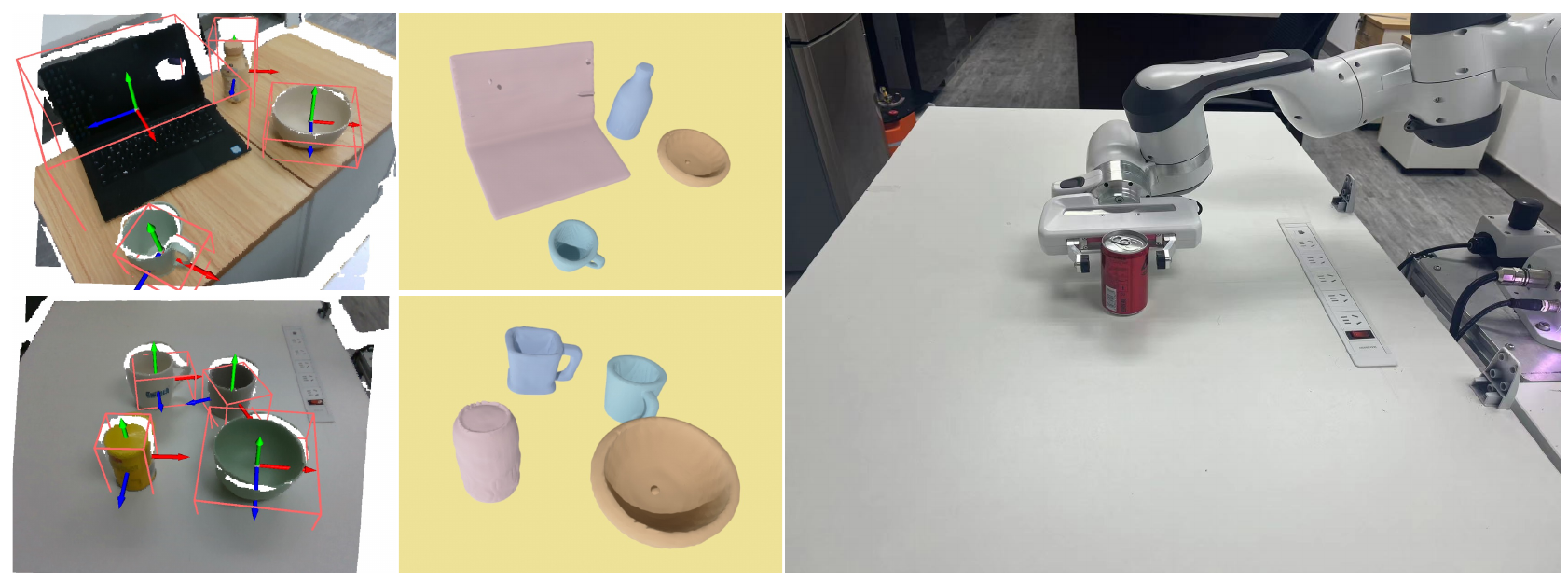}
  \caption{Robotic grasping experiments. A robot arm grasps an unseen object in the scene (right) with the help of {\dtf} for pose estimation (left) and shape reconstruction (middle).
  }
  \label{fig:real}
\end{figure}

To validate the effectiveness of our pose estimation and shape reconstruction, we also deployed the model trained on the REAL275 dataset~\cite{wang2019normalized} to control a real-world robotic platform with a 7-DoF robotic arm (Franka Emika Panda) for the grasping task, as shown in the right of Fig.\ref{fig:real}.
In the process of grasping, we followed the solution proposed by ShapePrior~\cite{tian2020shape} to map the ambiguous rotation to canonical rotation and then rotate the canonical rotation of the object 180 degrees counterclockwise along the x-axis to align with the coordinate system of the robotic gripper.
We evaluated three different shapes/colors/textures commonly found in life (\textit{bottle}, \textit{can}, \textit{cup}) and tried three grasping trials for each object, with more details and videos available in the supplementary materials.
Our method can accurately grasp the multi-object in the scene, and the grasping success rate is better than the baseline methods~\cite{shi2021fast,chen2021sgpa,irshad2022centersnap,di2022gpv}.

\section{Conclusion}
In this work, we propose a novel network for category-level 6D pose estimation and 3D reconstruction.
We design the deformable template fields that learn the template implicit field and predict the spatial deformation for unseen instances,
a object feature extraction module and a pose regression module to detect instances and estimate poses and 3D shapes,
and a implicit pose field and viewpoint sampling to enhance the pose representation.
Extensive experiments have demonstrated that our proposed method outperforms the state-of-the-art methods on the REAL275 and CAMERA25 datasets and is helpful in real-world grasping tasks.

\begin{acks}
  This research was supported in part by the National Key R\&D Program of China under Grant 2022YFF0904304,
the National Natural Science Foundation of China under Grant 62202065, 
the Project funded by China Postdoctoral Science Foundation 2022TQ0047 and 2022M710465, 
the BUPT Excellent Ph.D. Students Foundation under Grant CX2022224,
and the Science and Technology Innovation Action Plan of Shanghai under Grant 2251110540.
\end{acks}

\clearpage
\balance
\bibliographystyle{ACM-Reference-Format}
\bibliography{reference}

\end{document}